\def\year{2019}\relax
\documentclass[letterpaper]{article} 
\usepackage{aaai19}  
\usepackage{times}  
\usepackage{helvet}  
\usepackage{courier}  
\usepackage{url}  
\usepackage{graphicx,subfigure}  

\usepackage{color}
\usepackage{flushend}
\usepackage{amsfonts}
\usepackage{amssymb}
\usepackage{amsmath}
\usepackage[english]{babel}

\usepackage{algorithm,algorithmicx,algpseudocode}

\usepackage{enumitem}
\setenumerate[1]{itemsep=0pt,partopsep=0pt,parsep=\parskip,topsep=0pt}
\setitemize[1]{itemsep=0pt,partopsep=0pt,parsep=\parskip,topsep=0pt}
\setdescription{itemsep=0pt,partopsep=0pt,parsep=\parskip,topsep=0pt}

\begin{document}
%
\title{Detect or Track: Towards Cost-Effective Video Object Detection/Tracking}
\author{Hao Luo\textsuperscript{$1$}\thanks{This work was done when Hao Luo was an intern at Microsoft Research Asia.}, Wenxuan Xie\textsuperscript{$2$}, Xinggang Wang\textsuperscript{$1$}, Wenjun Zeng\textsuperscript{$2$}\\
\textsuperscript{$1$}School of Electronic Information and Communications, Huazhong University of Science and Technology\\
\textsuperscript{$2$}Microsoft Research Asia\\
\{luohao, xgwang\}@hust.edu.cn, \{wenxie, wezeng\}@microsoft.com
}
\maketitle
\begin{abstract}
  State-of-the-art object detectors and trackers are developing fast. Trackers are in general more efficient than detectors but bear the risk of drifting. A question is hence raised -- how to improve the accuracy of video object detection/tracking by utilizing the existing detectors and trackers within a given time budget? A baseline is frame skipping -- detecting every $N$-th frames and tracking for the frames in between. This baseline, however, is suboptimal since the detection frequency should depend on the tracking quality. To this end, we propose a scheduler network, which determines to detect or track at a certain frame, as a generalization of Siamese trackers. Although being light-weight and simple in structure, the scheduler network is more effective than the frame skipping baselines and flow-based approaches, as validated on ImageNet VID dataset in video object detection/tracking.
\end{abstract}

\section{Introduction}

Convolutional neural network (CNN)-based methods have achieved significant progress in computer vision tasks such as object detection \cite{ren2015faster,liu2016ssd,dai2016r,tang2018pcl} and tracking \cite{held2016learning,bertinetto2016fully,nam2016learning,bhat2018unveiling}. Following the tracking-by-detection paradigm, most state-of-the-art trackers can be viewed as a local detector of a specified object. Consequently, trackers are generally more efficient than detectors and can obtain precise bounding boxes in subsequent frames if the specified bounding box is accurate. However, as evaluated commonly on benchmark datasets such as OTB \cite{wu2015object} and VOT \cite{kristan2017visual}, trackers are encouraged to track as long as possible. It is non-trivial for trackers to be stopped once they are not confident, although heuristics, such as a threshold of the maximum response value, can be applied. Therefore, trackers bear the risk of drifting.

Besides object detection and tracking, there have been recently a series of studies on video object detection \cite{kang2016object,kang2017object,feichtenhofer2017detect,zhu2017deep,zhu2017flow,zhu2018towards,chen2018optimizing}. Beyond the baseline to detect each frame individually, state-of-the-art approaches consider the temporal consistency of the detection results via tubelet proposals \cite{kang2016object,kang2017object}, optical flow \cite{zhu2017deep,zhu2017flow,zhu2018towards} and regression-based trackers \cite{feichtenhofer2017detect}. These approaches, however, are optimized for the detection accuracy of each individual frame. They either do not associate the presence of an object in different frames as a tracklet, or associate after performing object detection on each frame, which is time-consuming.

This paper is motivated by the constraints from practical video analytics scenarios such as autonomous driving and video surveillance. We argue that algorithms applied to these scenarios should be:

\begin{itemize}
  \item capable of \textbf{associating an object} appearing in different frames, such that the trajectory or velocity of the object can be further inferred.
  \item in \textbf{realtime} (e.g., over 30 fps) and as fast as possible, such that the deployment cost can be further reduced.
  \item with \textbf{low latency}, which means to produce results once a frame in a video stream has been processed.
\end{itemize}

Considering these constraints, we focus in this paper on the task of video object detection/tracking \cite{russakovsky2017beyond}. The task is to detect objects in each frame (similar to the goal of video object detection), with an additional goal of associating an object appearing in different frames.

\begin{figure*}[!t]
  \centering
  \includegraphics[width=0.75\textwidth]{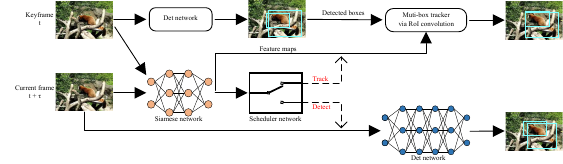}
  \caption{Detect or track (DorT) framework. The scheduler network compares the current frame $t+\tau$ with the keyframe $t$ by evaluating the tracking quality, and determines to \text{detect} or \textit{track} frame $t+\tau$: either frame $t+\tau$ is detected by a single-frame detector, or bounding boxes are tracked to frame $t+\tau$ from the keyframe $t$. If \textit{detect} is chosen, frame $t+\tau$ is assigned as the new keyframe, and the boxes in frame $t+\tau$ and frame $t+\tau-1$ are associated by the widely-used Hungarian algorithm (not shown in the figure for conciseness).}
  \label{fig:architecture}
\end{figure*}

In order to handle this task under the realtime and low latency constraint, we propose a detect or track (DorT) framework. In this framework, object detection/tracking of a video sequence is formulated as a sequential decision problem -- a scheduler network makes a detection/tracking decision for every incoming frame, and then these frames are processed with the detector/tracker accordingly. The architecture is illustrated in Figure \ref{fig:architecture}.

The scheduler network is the most unique part of our framework. It should be light-weight but be able to determine to detect or track. Rather than using heuristic rules (e.g., thresholds of tracking confidence values), we formulate the scheduler as a small CNN by assessing the tracking quality. It is shown to be a generalization of Siamese trackers and a special case of reinforcement learning (RL).

The contributions are summarized as follows:

\begin{itemize}
  \item We propose the DorT framework, in which the object detection/tracking of a video sequence is formulated as a sequential decision problem, while being in realtime and with low latency.
  \item We propose a light-weight but effective scheduler network, which is shown to be a generalization of Siamese trackers and a special case of RL.
  \item The proposed DorT framework is more effective than the frame skipping baselines and flow-based approaches, as validated on ImageNet VID dataset \cite{russakovsky2015imagenet} in video object detection/tracking.
\end{itemize}

\section{Related Work}

To our knowledge, we are the first to formulate video object detection/tracking as a sequential decision problem and there is no existing similar work to directly compare with. However, it is related to existing work in multiple aspects.

\subsection{Video Object Detection/Tracking}

Video object detection/tracking is a task in ILSVRC 2017 \cite{russakovsky2017beyond}, where the winning entries are optimized for accuracy rather than speed. \cite{deng2017speed} adopts flow aggregation \cite{zhu2017flow} to improve the detection accuracy. \cite{wei2017improving} combines flow-based \cite{ilg2017flownet} and object tracking-based \cite{nam2016learning} tubelet generation \cite{kang2017object}. THU-CAS \cite{russakovsky2017beyond} considers flow-based tracking \cite{kang2016object}, object tracking \cite{held2016learning} and data association \cite{yu2016poi}.

Nevertheless, these methods combine multiple cues (e.g., flow aggregation in detection, and flow-based and object tracking-based tubelet generation) which are complementary but time-consuming. Moreover, they apply global post-processing such as seq-NMS \cite{han2016seq} and tubelet NMS \cite{tang2018object} which greatly improve the accuracy but are not suitable for a realtime and low latency scenario.

\subsection{Video Object Detection}

Approaches to video object detection have been developed rapidly since the introduction of the ImageNet VID dataset \cite{russakovsky2015imagenet}. \cite{kang2016object,kang2017object} propose a framework that consists of per-frame proposal generation, bounding box tracking and tubelet re-scoring. \cite{zhu2017deep} proposes to detect frames sparsely and propagates features with optical flow. \cite{zhu2017flow} proposes to aggregate features in nearby frames along the motion path to improve the feature quality. Futhermore, \cite{zhu2018towards} proposes a high-performance approach by considering feature aggregation, partial feature updating and adaptive keyframe scheduling based on optical flow. Besides, \cite{feichtenhofer2017detect} proposes to learn detection and tracking using a single network with a multi-task objective. \cite{chen2018optimizing} proposes to propagate the sparsely detected results through a space-time lattice. All the methods above focus on the accuracy of each individual frame. They either do not associate the presence of an object in different frames as a tracklet, or associate after performing object detection on each frame, which is time-consuming.

\subsection{Multiple Object Tracking}

Multiple object tracking (MOT) focuses on data association: finding the set of trajectories that best explains the given detections \cite{leal2014learning}. Existing approaches to MOT fall into two categories: batch and online mode. Batch mode approaches pose data association as a global optimization problem, which can be a min-cost max-flow problem \cite{zhang2008global,pirsiavash2011globally}, a continuous energy minimization problem \cite{milan2014continuous} or a graph cut problem \cite{tang2016multi,tang2017multiple}. Contrarily, online mode approaches are only allowed to solve the data association problem with the present and past frames. \cite{xiang2015learning} formulates data association as a Markov decision process. \cite{milan2017online,sadeghian2017tracking} employs recurrent neural networks (RNNs) for feature representation and data association.

State-of-the-art MOT approaches aim to improve the data association performance given publicly-available detections since the introduction of the MOT challenge \cite{leal2015motchallenge}. However, we focus on the sequential decision problem of detection or tracking. Although the widely-used Hungarian algorithm is adopted for simplicity and fairness in the experiments, we believe the incorporation of existing MOT approaches can further enhance the accuracy.

\subsection{Keyframe Scheduler}

Researchers have proposed approaches to adaptive keyframe scheduling beyond regular frame skipping in video analytics. \cite{zhu2018towards} proposes to estimate the quality of optical flow, which relies on the time-consuming flow network. \cite{chen2018optimizing} proposes an \textit{easiness measure} to consider the size and motion of small objects, which is hand-crafted and more importantly, it is a detect-then-schedule paradigm but cannot determine to detect or track. \cite{li2018low,xu2018dynamic} learn to predict the discrepancy between the segmentation map of the current frame and the keyframe, which are only applicable to segmentation tasks.

All the methods above, however, solve an auxiliary task (e.g., flow quality, or discrepancy of segmentation maps) but do not answer the question directly in a classification perspective -- is the current frame a keyframe or not? In contrast, we pose video object detection/tracking as a sequential decision problem, and learn directly whether the current frame is a keyframe by assessing the tracking quality. Our formulation is further shown as a generalization of Siamese trackers and a special case of RL.

\section{The DorT Framework}

Video object detection/tracking is formulated as follows. Given a sequence of video frames $F=\{f_1,f_2,\ldots,f_N\}$, the aim is to obtain bounding boxes $B=\{b_1,b_2,\ldots,b_M\}$, where $b_i=\{rect_i,fid_i,score_i,id_i\}$, $rect_i$ denotes the 4-dim bounding box coordinates and $fid_i$, $score_i$ and $id_i$ are scalars denoting respectively the frame ID, the confidence score and the object ID.

Considering the realtime and low latency constraint, we formulate video object detection/tracking as a sequential decision problem, which consists of four modules: single-frame detector, multi-box tracker, scheduler network and data association. An algorithm summary follows the introduction of the four modules.

\subsection{Single-Frame Detector}

We adopt R-FCN \cite{dai2016r} as the detector following deep feature flow (DFF) \cite{zhu2017deep}. Our framework, however, is compatible with all single-frame detectors.

\subsection{Efficient Multi-Box Tracker via RoI Convolution}

The SiamFC tracker \cite{bertinetto2016fully} is adopted in our framework. It learns a deep feature extractor during training such that an object is similar to its deformations but different from the background. During testing, the nearby patch with the highest confidence is selected as the tracking result. The tracker is reported to run at 86 fps in the original paper.

Despite its efficiency, there are usually 30 to 50 detected boxes in a frame outputted by R-FCN. It is a natural idea to track only the high-confidence ones and ignore the rest. Such an approach, however, results in a drastic decrease in mAP since R-FCN detection is not perfect and many true positives with low confidence scores are discarded. We therefore need to track all the detected boxes.

\begin{figure}[!t]
  \centering
  \includegraphics[width=0.8\columnwidth]{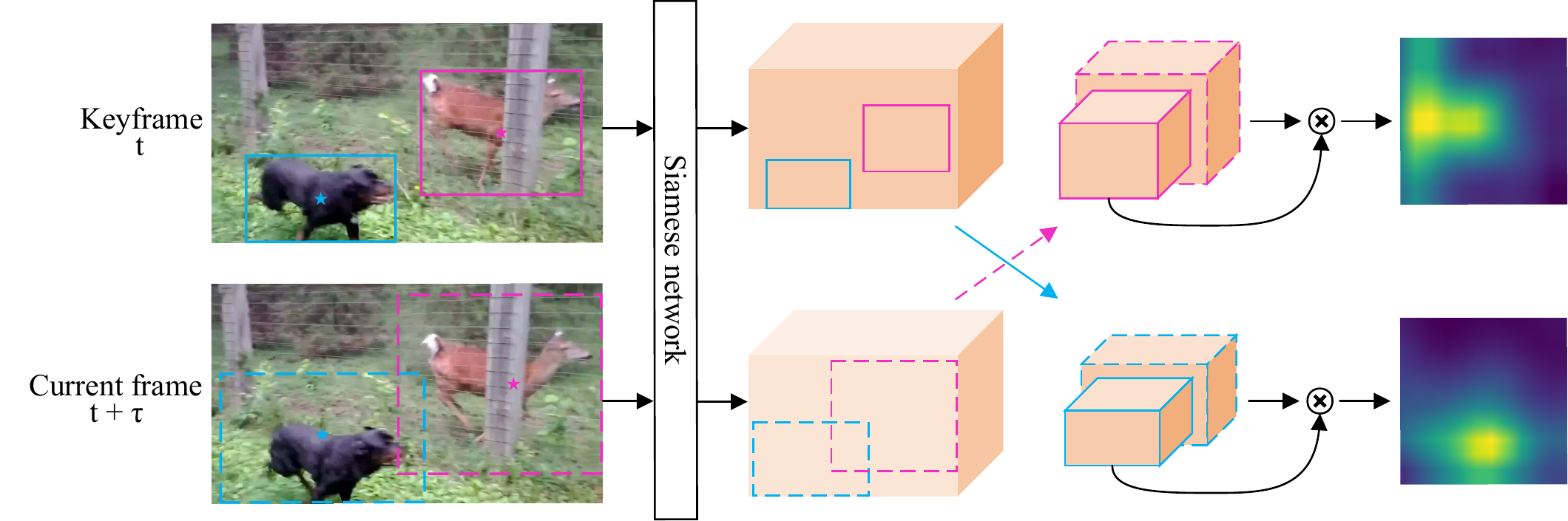}
  \caption{RoI convolution. Given targets in keyframe $t$ and search regions in frame $t+\tau$, the corresponding RoIs are cropped from the feature maps and convolved to obtain the response maps. Solid boxes denote detected objects in keyframe $t$ and dashed boxes denote the corresponding search region in frame $t+\tau$. A star $\star$ denotes the center of its corresponding bounding box. The center of a dashed box is copied from the tracking result in frame $t+\tau-1$.}
  \label{fig:roi-convolution}
\end{figure}

It is time-consuming to track 50 boxes without optimization (about 3 fps). In order to speed up the tracking process, we propose to share the feature extraction network of multiple boxes and propose an RoI convolution layer in place of the original cross-correlation layer in SiamFC. Figure \ref{fig:roi-convolution} is an illustration. Through cropping and convolving on the feature maps, the proposed tracker is over 10x faster than the time-consuming baseline while obtaining comparable accuracy.

Notably, there is no learnable parameter in the RoI convolution layer, and thus we can train the SiamFC tracker following the original settings in \cite{bertinetto2016fully}.

\subsection{Scheduler Network}

The scheduler network is the core of DorT, as our task is formulated as a sequential decision problem. It takes as input the current frame $f_{t+\tau}$ and its keyframe $f_t$, and determines to detect or track, denoted as $Scheduler(f_t,f_{t+\tau})$. We will elaborate this module in the next section.

\subsection{Data Association}

Once the scheduler network determines to detect the current frame, there is a need to associate the previous tracked boxes and the current detected boxes. Hence, a data association algorithm is required. For simplicity and fairness in the paper, the widely-used Hungarian algorithm is adopted. Although it is possible to improve the accuracy by incorporating more advanced data association techniques \cite{xiang2015learning,sadeghian2017tracking}, it is not the focus in the paper. The overall architecture of the DorT framework is shown in Figure \ref{fig:architecture}. More details are summarized in Algorithm \ref{alg:DorT-framework}.

\begin{algorithm}[!t]
  \small
  \caption{The Detect or Track (DorT) Framework}
  \label{alg:DorT-framework}
  \begin{algorithmic}[1]
  \Require A sequence of video frames $F=\{f_1,f_2,\ldots,f_N\}$.
  \Ensure Bounding boxes $B=\{b_1,b_2,\ldots,b_M\}$ with ID, where $b_i=\{rect_i,fid_i,score_i,id_i\}$.
  \State{$B\gets \{\}$}
  \State{$t\gets 1$} \Comment{$t$ is the index of keyframe}
  \State{Detect $f_1$ with the single-frame detector.}
  \State{Assign new ID to the detected boxes.}
  \State{Add the detected boxes in $f_1$ to $B$.}
  \For{$i\gets 2$ to $N$}
    \State{$d\gets Scheduler(f_t,f_i)$} \Comment{decision of scheduler}
    \If{$d=detect$}
      \State{Detect $f_i$ with single-frame detector.}
      \State{Match boxes in $f_i$ and $f_{i-1}$ using Hungarian.}
      \State{Assign new ID to unmatched boxes in $f_i$.}
      \State{Assign corresponding ID to matched boxes in $f_i$.}
      \State{$t\gets i$} \Comment{update keyframe}
    \Else \Comment{the decision is to $track$}
      \State{Track boxes from $f_t$ to $f_i$.}
      \State{Assign corresponding ID to tracked boxes in $f_i$.}
      \State{Assign corresponding detection score to tracked boxes in $f_i$.}
    \EndIf
    \State{Add the bounding boxes in $f_i$ to $B$.}
  \EndFor
  \end{algorithmic}
\end{algorithm}

\section{The Scheduler Network in DorT}

\begin{figure}[!t]
  \centering
  \includegraphics[width=0.8\columnwidth]{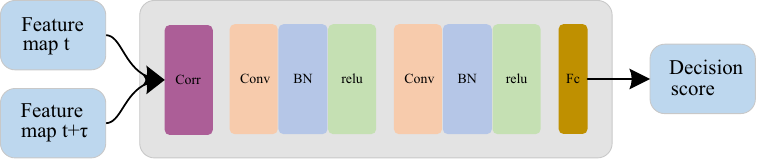}
  \caption{Scheduler network. The output feature map of the correlation layer is followed by two convolutional layers and a fully-connected layer with a 2-way softmax. As discussed later, this structure is a generalization of the SiamFC tracker.}
  \label{fig:scheduler}
\end{figure}

The scheduler network in DorT aims to determine to detect or track given a new frame by estimating the quality of the tracked boxes. It should be efficient itself. Rather than training a network from scratch, we propose to reuse part of the tracking network. Firstly, the $l$-th layer convolutional feature map of frame $t$ and frame $t+\tau$, denoted respectively as $x_l^t$ and $x_l^{t+\tau}$, are fed into a correlation layer which performs point-wise feature comparison
\begin{equation}
x_{corr}^{t,t+\tau}(i,j,p,q)=\Big<x_l^t(i,j),x_l^{t+\tau}(i+p,j+q)\Big>
\label{eq:correlation}
\end{equation}
where $-d\leq p\leq d$ and $-d\leq q\leq d$ are offsets to compare features in a neighbourhood around the locations $(i,j)$ in the feature map, defined by the maximum displacement $d$. Hence, the output of the correlation layer is a feature map of size $x_{corr}\in \mathbb{R}^{H_l\times W_l\times (2d+1)^2}$, where $H_l$ and $W_l$ denote respectively the height and width of the $l$-th layer feature map. The correlation feature map $x_{corr}$ is then passed through two convolutional layers and a fully-connected layer with a 2-way softmax. The final output of the network is a classification score indicating the probability to detect the current frame. Figure \ref{fig:scheduler} is an illustration of the scheduler network.

\subsection{Training Data Preparation}

Existing groundtruth in the ImageNet VID dataset \cite{russakovsky2015imagenet} does not contain an indicator of the tracking quality. In this paper, we simulate the tracking process between two sampled frames and label it as \textit{detect} (0) or \textit{track} (1) in a strict protocol.

\begin{figure*}[!t]
  \centering
  \subfigure[Positive examples]{\includegraphics[width=0.4\textwidth]{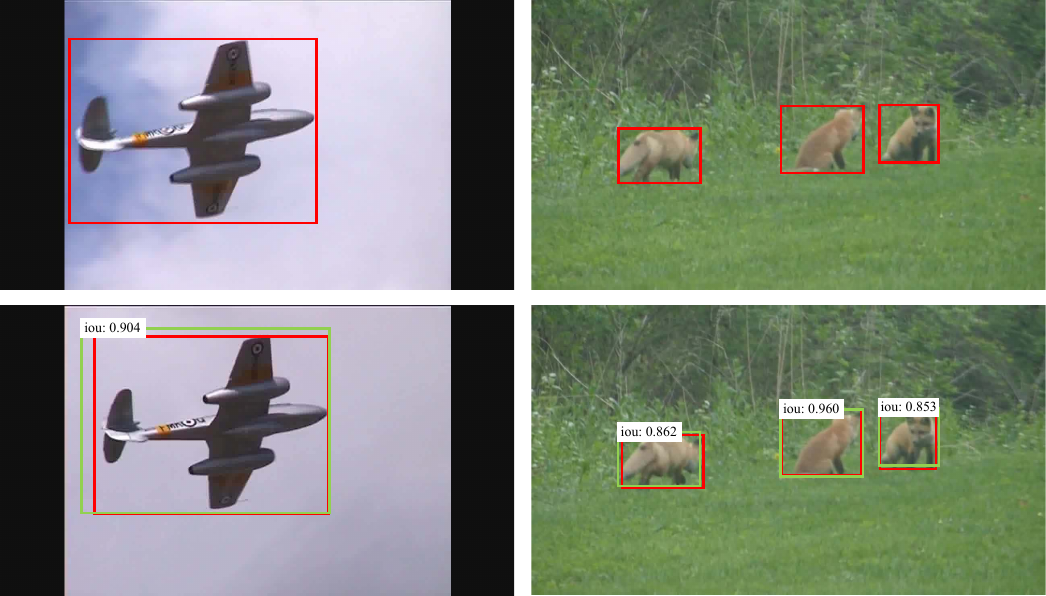}}
  \subfigure[Negative examples]{\includegraphics[width=0.4\textwidth]{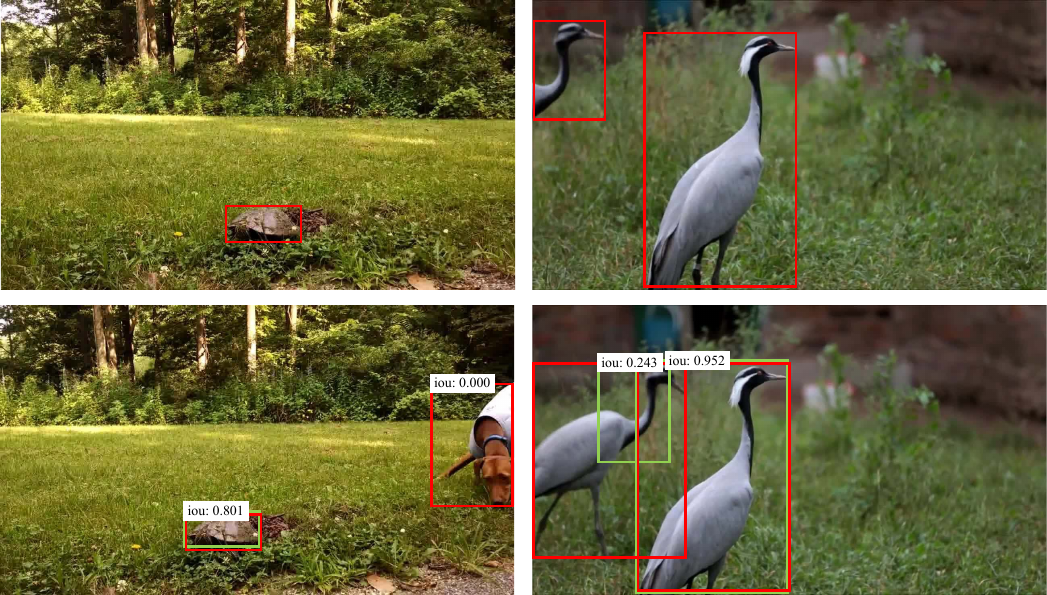}}
  \caption{Examples of labeled data for training the scheduler network. \textcolor{red}{Red} and \textcolor{green}{green} boxes denote groundtruth and tracked results, respectively. (a) Positive examples, where the IOU of each groundtruth box and its corresponding tracked box is over a threshold; (b) Negative examples, where at least one such IOU is below a threshold or there are emerging/disappearing  objects.}
  \label{fig:labeled-examples}
\end{figure*}

As we have sampled frame $t$ and frame $t+\tau$ from the same sequence, we track all the groundtruth bounding boxes using SiamFC from frame $t$ to frame $t+\tau$. If all the groundtruth boxes in frame $t+\tau$ are matched with the tracked boxes (e.g., IOU over $0.8$), the frame is labeled as \textit{track}; otherwise, it is labeled as \textit{detect}. Any emerging or disappearing objects indicates a \textit{detect}. Several examples are shown in Figure \ref{fig:labeled-examples}.

We have also tried to learn a scheduler for each tracker, but found it difficult to handle high-confidence false detections and non-trivial to merge the decisions of all the trackers. In contrast, the proposed approach to learning a single scheduler is an elegant solution which directly learns the decision rather than an auxiliary target such as the fraction of pixels at which the semantic segmentation labels differ \cite{li2018low}, or the fraction of low-quality flow estimation \cite{zhu2018towards}.

\subsection{Relation to the SiamFC Tracker}

\begin{figure}[!t]
  \centering
  \includegraphics[width=0.8\columnwidth]{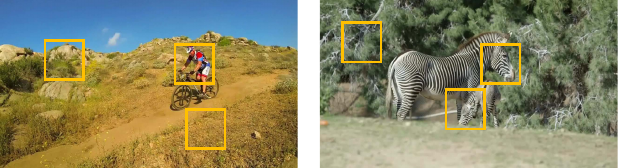}
  \caption{Examples of targets on keyframes. The size of a target equals the receptive field of the input feature map of the scheduler. As shown, a target patch might be an object, a part of an object, or totally background. The ``tracking'' results of these targets will be fused later. It should be noted that targets include all possible patches in a sliding window manner, but not just the three boxes shown above.}
  \label{fig:target-examples}
\end{figure}

The proposed scheduler network can be seen as a generalization of the original SiamFC \cite{bertinetto2016fully}. In the correlation layer of SiamFC, the target feature ($6\times 6\times 128$) is convolved with the search region feature ($22\times 22\times 128$) and obtains the response map ($17\times 17\times 1$, which can be equivalently written as $1\times 1\times 17^2$). Similarly, we can view the correlation layer of the proposed scheduler network (see Eq. \ref{eq:correlation}) as convolutions between multiple target features in the keyframe and their respective nearby search regions in the current frame. The size of a target equals the receptive field of the input feature map of our scheduler. Figure \ref{fig:target-examples} shows several examples of targets. Actually, however, targets include all possible patches in a sliding window manner.

In this sense, the output feature map of the correlation layer $x_{corr}\in \mathbb{R}^{H_l\times W_l\times (2d+1)^2}$ can be regarded as a set of $H_l\times W_l$ SiamFC tracking tasks, where the response map of each is $1\times 1\times (2d+1)^2$. The correlation feature map is then fed into a small CNN consisting of two convolutional layers and a fully-connected layer.

In summary, the generalization of the proposed scheduler network over SiamFC lies in two fold:

\begin{itemize}
  \item SiamFC correlates a target feature with its nearby search region, while our scheduler extends the number of tasks from one to many.
  \item SiamFC directly picks the highest value in the correlation feature map as the result, whereas the proposed scheduler fuses the multiple response maps with a CNN.
\end{itemize}

The validity of the proposed scheduler network is hence clear -- it first convolves patches in frame $t$ (examples shown in Figure \ref{fig:target-examples}) with their respective nearby regions in frame $t+\tau$, and then fuses the response maps with a CNN, in order to measure the difference between the two frames, and more importantly, to assess the tracking quality. The scheduler is also resistant to small perturbations by inheriting SiamFC's robustness to object deformation.

\subsection{Relation to Reinforcement Learning}

The sequential decision problem can also be formulated in a RL framework, where the action, state, state transition function and reward need to be defined.

\subsubsection{Action.}

The action space $\mathcal{A}$ contains two types of actions: \{\textit{detect}, \textit{track}\}. If the decision is \textit{detect}, object detector is applied to the current frame; otherwise, boxes tracked from the keyframe are taken as the results.

\subsubsection{State.}

The state $s_{t, \tau}$ is defined as a tuple $(x_l^t, x_l^{t+\tau})$, where $x_l^t$ and $x_l^{t+\tau}$ denote the $l$-th layer convolutional feature map of frame $t$ and frame $t+\tau$, respectively. Frame $t$ is the keyframe on which object detector is applied, and frame $t+\tau$ is the current frame on which actions are to be determined.

\subsubsection{State transition function.}

After the decision of action $a_{t, \tau}$ in state $s_{t, \tau}$. The next state is obtained upon the action:

\begin{itemize}
  \item \textit{detect}. The next state is $s_{t+\tau, 1}=(x_l^{t+\tau}, x_l^{t+\tau+1})$. Frame $t+\tau$ is fed to the object detector and is set as the new keyframe.
  \item \textit{track}. The next state is $s_{t, \tau+1}=(x_l^t, x_l^{t+\tau+1})$. Bounding boxes tracked from the keyframe are taken as the results in frame $t+\tau$. The keyframe $t$ remains unchanged.
\end{itemize}

As shown above, no matter whether the keyframe is $t$ or $t+\tau$, the task in the next state is to determine the action in frame $t+\tau+1$.

\subsubsection{Reward.}

The reward function is defined as $r(s,a)$ since it is determined by both the state $s$ and the action $a$. As illustrated in Figure \ref{fig:labeled-examples}, a labeling mechanism is proposed to obtain the groundtruth label of the tracking quality between two frames (i.e., a certain state $s$). We denote the groundtruth label as $GT(s)$, which is either \textit{detect} or \textit{track}. Hence, the reward function can be defined as follows:
\begin{equation}
r(s,a)=
\begin{cases}
1,& GT(s)=a\\
0,& GT(s)\neq a
\end{cases}
\end{equation}
which is based on the consistency between the groundtruth label and the action taken.

After defining all the above, the RL problem can be solved via a deep Q network (DQN) \cite{mnih2015human} with a discount factor $\gamma$, penalizing the reward from future time steps. However, training stability is always an issue in RL algorithms \cite{anschel2016averaged}. In this paper, we set $\gamma=0$ such that the agent only cares about the reward from the next time step. Therefore, the DQN becomes a regression network -- pushing the predicted action to be the same as the \textit{GT} action, and the scheduler network is a special case of RL. We empirically observe that the training procedure becomes easier and more stable by setting $\gamma=0$.

\section{Experiments}

The DorT framework is evaluated on the ImageNet VID dataset \cite{russakovsky2015imagenet} in the task of video object detection/tracking. For completeness, we also report results in video object detection.

\subsection{Experimental Setup}

\subsubsection{Dataset description.}

All experiments are conducted on the ImageNet VID dataset \cite{russakovsky2015imagenet}. ImageNet VID is split into a training set of 3862 videos and a test set of 555 videos. There are per-frame bounding box annotations for each video. Furthermore, the presences of a certain target across different frames in a video are assigned with the same ID.

\subsubsection{Evaluation metric.}

The evaluation metric for video object detection is the extensively used mean average precision (mAP), which is based on a sorted list of bounding boxes in descending order of their scores. A predicted bounding box is considered correct if its IOU with a groundtruth box is over a threshold (e.g., $0.5$).

In contrast to the standard mAP which is based on bounding boxes, the mAP for video object detection/tracking is based on a sorted list of tracklets \cite{russakovsky2017beyond}. A tracklet is a set of bounding boxes with the same ID. Similarly, a tracklet is considered correct if its IOU with a groundtruth tracklet is over a threshold. Typical choices of IOU thresholds for tracklet matching and per-frame bounding box matching are both $0.5$. The score of a tracklet is the average score of all its bounding boxes.

\subsubsection{Implementation details.}

Following the settings in \cite{zhu2017deep}, R-FCN \cite{dai2016r} is trained with a ResNet-101 backbone \cite{he2016deep} on the training set.

SiamFC is trained following the original paper \cite{bertinetto2016fully}. Instead of training from scratch, however, we initialize the first four convolutional layers with the pretrained parameters from AlexNet \cite{krizhevsky2012imagenet} and change Conv5 from $3\times 3$ to $1\times 1$ with the Xavier initializer. Parameters of the first four convolutional layers are fixed during training \cite{he2018twofold}. We only search for one scale and discard the upsampling step in the original SiamFC for efficiency. All images being fed into SiamFC are resized to $300\times 500$. Moreover, the confidence score of a tracked box (for evaluation) is equal to its corresponding detected box in the keyframe.

The scheduler network takes as input the Conv5 feature of our trained SiamFC. The SGD optimizer is adopted with a learning rate 1e-2, momentum 0.9 and weight decay 5e-4. The batch size is set to 32. During testing, we raise the decision threshold of \textit{track} to $\delta=0.97$ (i.e., the scheduler outputs \textit{track} if the predicted confidence of \textit{track} is over $\delta$) to ensure conservativeness of the scheduler. Furthermore, since nearby frames look similar, the scheduler is applied every $\sigma$ frames (where $\sigma$ is a tunable parameter) to reduce unnecessary computation.

All experiments are conducted on a workstation with an Intel Core i7-4790k CPU and a Titan X GPU. We empirically observe that the detection network and the tracking/scheduler network run at 8.33 fps and 100fps, respectively. This is because the ResNet-101 backbone is much heavier than AlexNet. Moreover, the speed of the Hungarian algorithm is as high as 667 fps.

\subsection{Video Object Detection/Tracking}

To our knowledge, the most closely related work to ours is \cite{lan2016super}, which handles cost-effective face detection/tracking. Since face is much easier to track and is with less deformation, the paper achieves success by utilizing non-deep learning-based detectors and trackers. However, we aim at general object detection/tracking in video, which is much more challenging. We demonstrate the effectiveness of the proposed DorT framework against several strong baselines.

\subsubsection{Effectiveness of scheduler.}

The scheduler network is a core component of our DorT framework. Since SiamFC tracking is more efficient than R-FCN detection, the scheduler should predict \textit{track} when it is safe for the trackers and be conservative enough to predict \textit{detect} when there is sufficient change to avoid track drift.

\begin{figure}[!t]
  \centering
  \includegraphics[width=0.7\columnwidth]{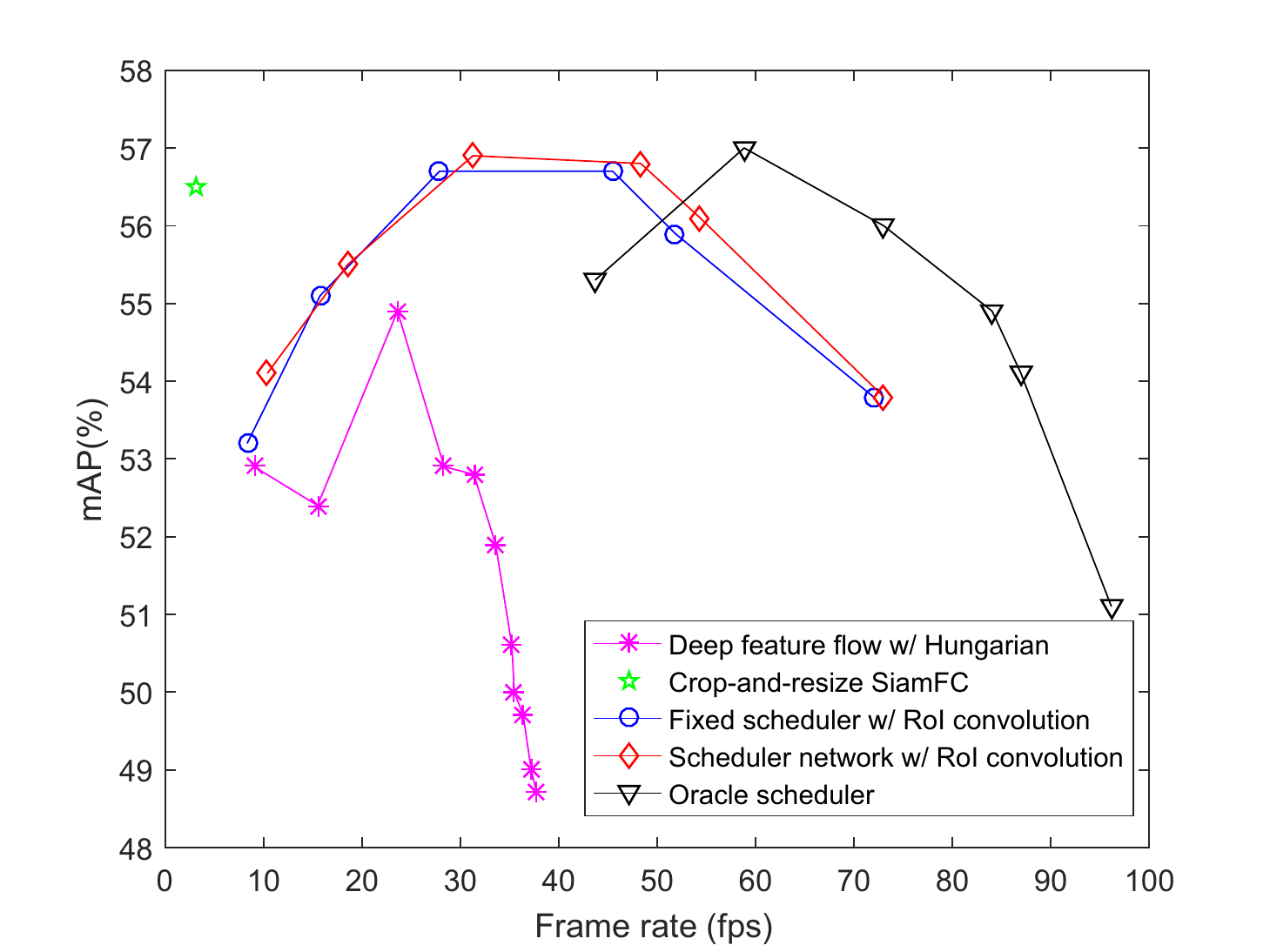}
  \caption{Comparison between different methods in video object detection/tracking in terms of mAP. The detector (for deep feature flow and fixed scheduler) or the scheduler (for scheduler network and oracle scheduler) can be applied every $\sigma$ frames to obtain different results.}
  \label{fig:vodt-map}
\end{figure}

We compare our DorT framework with a frame skipping baseline, namely a ``fixed scheduler'' -- R-FCN is performed every $\sigma$ frames and SiamFC is adopted to track for the frames in between. As aforementioned, our scheduler can also be applied every $\sigma$ frames to improve efficiency. Moreover, there could be an oracle scheduler -- predicting the groundtruth label (\textit{detect} or \textit{track}) as shown in Figure \ref{fig:labeled-examples} during testing. The oracle scheduler is a 100\% accurate scheduler in our setting. The results are shown in Figure \ref{fig:vodt-map}.

We can observe that the frame rate and mAP vary as $\sigma$ changes. Interestingly, the curves are not monotonic -- as the frame rate decreases, the accuracy in mAP is not necessarily higher. In particular, detectors are applied frequently when $\sigma=1$ (the leftmost point of each curve). Associating boxes using the Hungarian algorithm is generally less reliable (given missed detections and false detections) than tracking boxes between two frames. It is also a benefit of the scheduler network -- applying tracking only when confident, and thus most boxes are reliably associated. Hence, the curve of the scheduler network is on the upper-right side of that of the fixed scheduler as shown in Figure \ref{fig:vodt-map}.

\begin{figure}[!t]
  \centering
  \subfigure[$\sigma=1$]{\includegraphics[width=0.4\columnwidth]{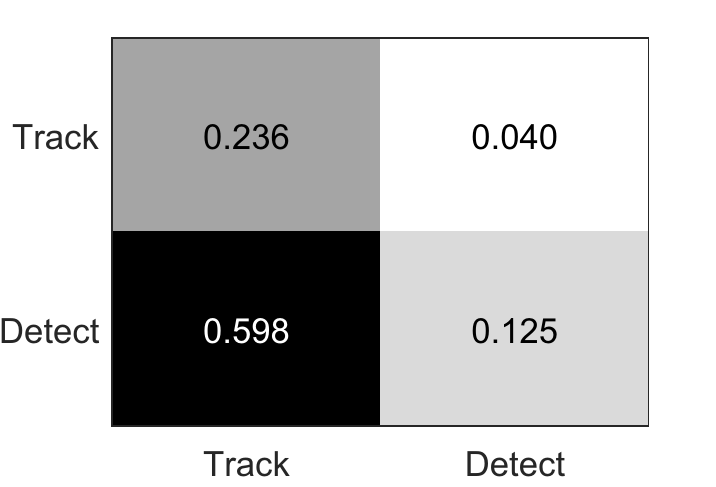}}
  \subfigure[$\sigma=10$]{\includegraphics[width=0.4\columnwidth]{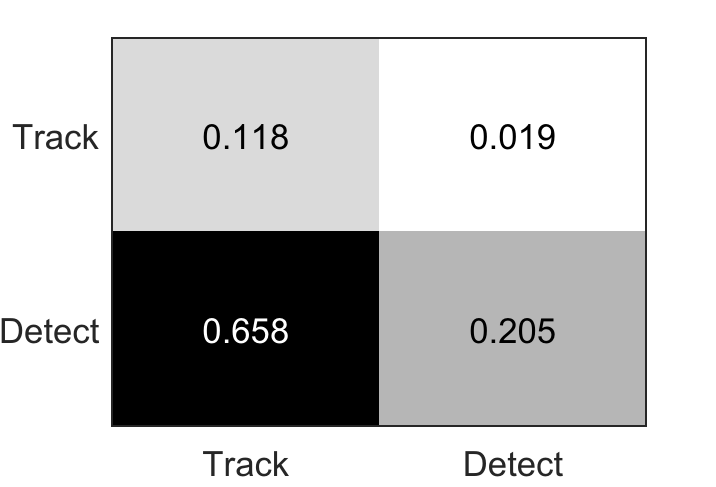}}
  \caption{Confusion matrix of the scheduler network. The horizontal axis is the groundtruth and the vertical axis is the predicted label. The scheduler is applied every $\sigma$ frames.}
  \label{fig:confusion}
\end{figure}

\begin{figure}[!t]
  \centering
  \includegraphics[width=0.8\columnwidth]{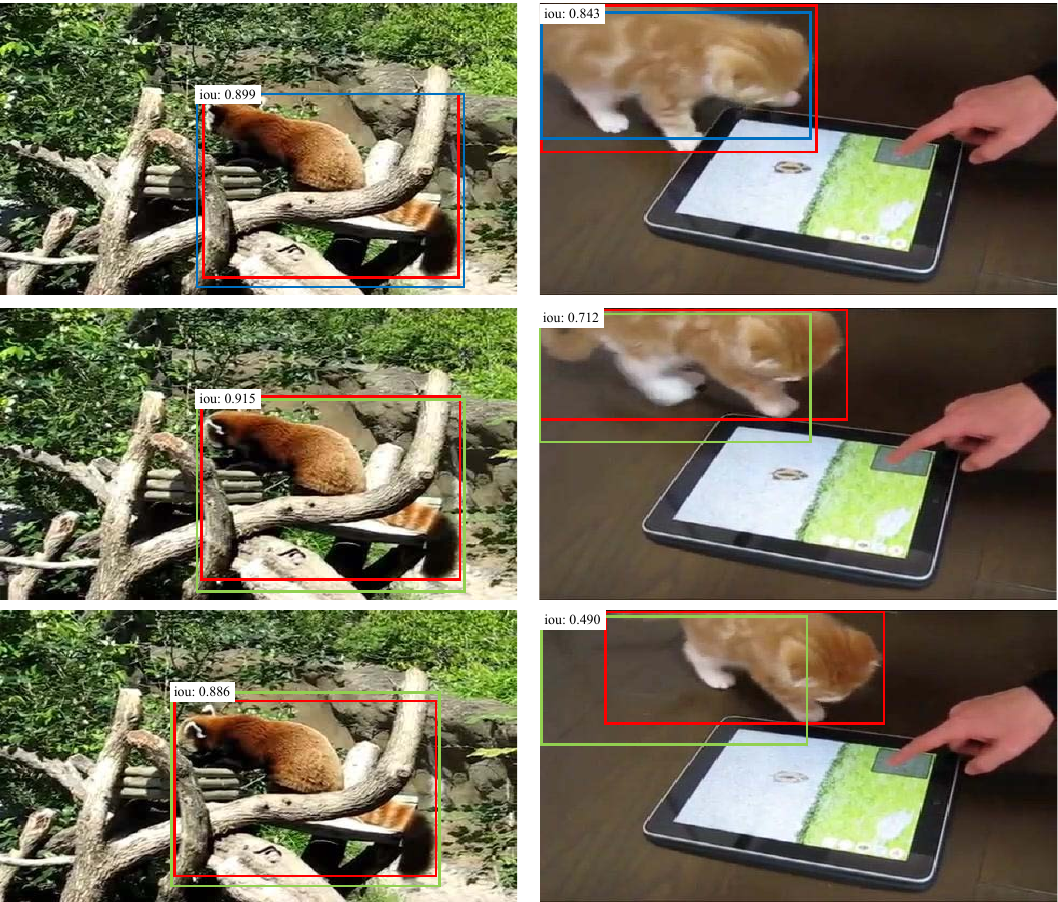}
  \caption{Qualitative results of the scheduler network. \textcolor{red}{Red}, \textcolor{blue}{blue} and \textcolor{green}{green} boxes denote groundtruth, detected boxes and tracked boxes, respectively. The first row: R-FCN is applied in the keyframe. The second row: the scheduler determines to \textit{track} since it is confident. The third row: the scheduler predicts to \textit{track} in the first image although the red panda moves; however, the scheduler determines to \textit{detect} in the second image as the cat moves significantly and is unable to be tracked.}
  \label{fig:qualitative-results}
\end{figure}

However, it can be also observed that there is certain distance between the curve of the scheduler network and that of the oracle scheduler. Given that the oracle scheduler is a 100\% accurate classifier, we analyze the classification accuracy of the scheduler network in Figure \ref{fig:confusion}.
Let us take the $\sigma=10$ case as an example. Although the classification accuracy is only 32.3\%, the false positive rate (i.e., misclassifying a \textit{detect} case as \textit{track}) is as low as 1.9\%. Because we empirically find that the mAP drops drastically if the scheduler mistakenly predicts \textit{track}, our scheduler network is made conservative -- \textit{track} only when confident and \textit{detect} if unsure. Figure \ref{fig:qualitative-results} shows some qualitative results.

\subsubsection{Effectiveness of RoI convolution.}

Trackers are optimized for the crop-and-resize case \cite{bertinetto2016fully} -- the target and search region are cropped and resized to a fixed size before matching. It is a nice choice since the tracking algorithm is not affected by the original size of the target. It is, however, slow in multi-box case and we propose RoI convolution as an efficient approximation. As shown in Figure \ref{fig:vodt-map}, crop-and-resize SiamFC is even slower than detection -- the overall running time is 3 fps. Notably, its mAP is 56.5\%, which is roughly the same as that of our DorT framework empowered with RoI convolution. Our DorT framework, however, runs at 54 fps when $\sigma=10$. RoI convolution obtains over 10x speed boost while retaining mAP.

\subsubsection{Comparison with existing methods.}

Deep feature flow \cite{zhu2017deep} focuses on video object detection without tracking. We can, however, associate its predicted bounding boxes with per frame data association using the Hungarian algorithm. The results are shown in Figure \ref{fig:vodt-map}. It can be observed that our framework performs significantly better than deep feature flow in video object detection/tracking.

Concurrent works that deal with video object detection/tracking are the submitted entries in ILSVRC 2017 \cite{deng2017speed,wei2017improving,russakovsky2017beyond}. As discussed in the Related Work section, these methods aim only to improve the mAP by adopting complicated methods and post processing, leading to inefficient solutions without guaranteeing low latency. Their reported results on the test set ranges from 51\% to 65\% mAP. Our proposed DorT, notably, achieves 57\% mAP on the validation set, which is comparable to the existing methods in magnitude, but is much more principled and efficient.

\subsection{Video Object Detection}

\begin{figure}[!t]
  \centering
  \includegraphics[width=0.7\columnwidth]{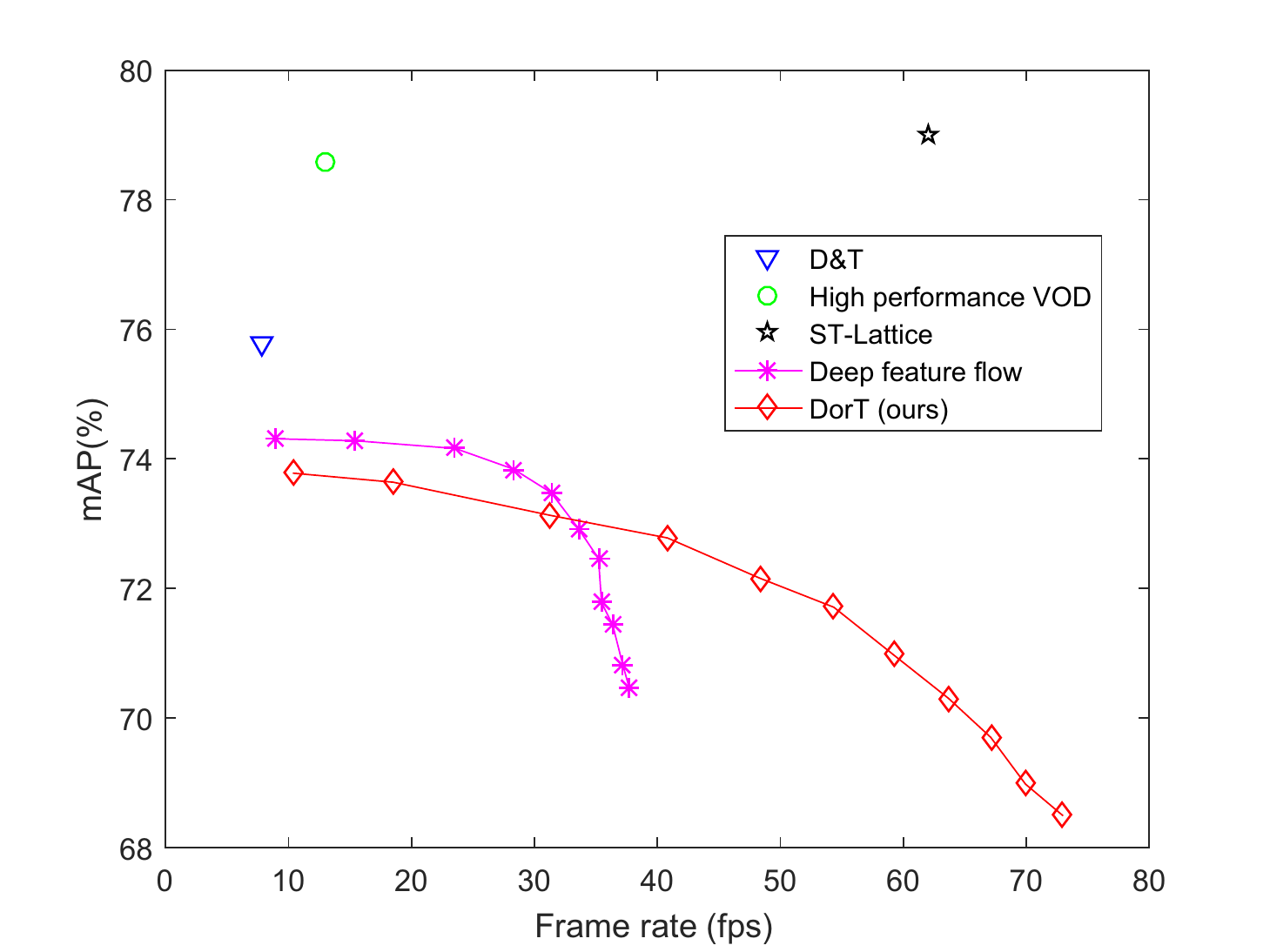}
  \caption{Comparison between different methods in video object detection in terms of mAP. Results of D\&T, High performance VOD and ST-Lattice are copied from the original papers. The detector (for deep feature flow) or the scheduler (for scheduler network) can be applied every $\sigma$ frames to obtain different results.}
  \label{fig:vod-map}
\end{figure}

We also evaluate our DorT framework in video object detection for completeness, by removing the predicted object ID. Our DorT framework is compared against deep feature flow \cite{zhu2017deep}, D\&T \cite{feichtenhofer2017detect}, high performance video object detection (VOD) \cite{zhu2018towards} and ST-Lattice \cite{chen2018optimizing}. The results are shown in Figure \ref{fig:vod-map}.
It can be observed that D\&T and high performance VOD manage to achieve a speed-accuracy balance. They obtain higher results but cannot fit into realtime (over 30 fps) scenarios. ST-Lattice, although being fast and accurate, adopts detection results in future frames and is thus not suitable in a low latency scenario. As compared with deep feature flow, our DorT framework performs significantly faster with comparable performance (no more than 1\% mAP loss). Although our aim is not the video object detection task, the results in Figure \ref{fig:vod-map} demonstrate the effectiveness of our approach.

\section{Conclusion and Future Work}

We propose a DorT framework for cost-effective video object detection/tracking, which is in realtime and with low latency. Object detection/tracking of a video sequence is formulated as a sequential decision problem in the framework. Notably, a light-weight but effective scheduler network is proposed, which is shown to be a generalization of Siamese trackers and a special case of RL. The DorT framework turns out to be effective and strikes a good balance between speed and accuracy.

The framework can still be improved in several aspects. The SiamFC tracker can search for multiple scales to improve performance as in the original paper. More advanced data association methods can be applied by resorting to the state-of-the-art MOT algorithms. Furthermore, there is room to improve the training of the scheduler network to approach the oracle scheduler. These are left as future work.

\subsubsection{Acknowledgment.} This work was partly supported by NSFC (No. 61876212 \& 61733007). The authors would like to thank Chong Luo and Anfeng He for fruitful discussions.

\bibliographystyle{aaai}

\end{document}